\documentclass[conference]{IEEEtran}
\usepackage{titlesec}
\usepackage{adjustbox}
\usepackage [english]{babel}
\usepackage [autostyle, english = american]{csquotes}
\MakeOuterQuote{"}
\usepackage{hyperref}
\usepackage{algorithm}
\usepackage{algorithmic}
\usepackage{graphicx}
\titleformat{\paragraph}
{\normalfont\normalsize\bfseries}{\theparagraph}{1em}{}
\titlespacing*{\paragraph}
{0pt}{3.25ex plus 1ex minus .2ex}{1.5ex plus .2ex}
\IEEEoverridecommandlockouts
\usepackage{cite}
\usepackage{amsmath,amssymb,amsfonts}
\usepackage{algorithmic}
\usepackage{graphicx}
\usepackage{hyperref}
\usepackage{textcomp}
\usepackage{makecell}
\usepackage{multirow,bigdelim,dcolumn,booktabs}
\usepackage{array}
\usepackage{float}
\usepackage{enumitem} 
\usepackage[caption=false]{subfig}
\usepackage{url}
\usepackage{xurl}
\usepackage{multirow, booktabs}
\usepackage{etoolbox}
\makeatletter
\patchcmd{\@makecaption}
  {\scshape}
  {}
  {}
  {}
\makeatother

\usepackage{xcolor}
\def\BibTeX{{\rm B\kern-.05em{\sc i\kern-.025em b}\kern-.08em
    T\kern-.1667em\lower.7ex\hbox{E}\kern-.125emX}}
\begin{document}

\title{Advanced Volleyball Stats for All Levels: Automatic Setting Tactic Detection and Classification with a Single Camera\\
}

\author{\IEEEauthorblockN{Haotian Xia\IEEEauthorrefmark{2},
Rhys Tracy\IEEEauthorrefmark{3}, Yun Zhao\IEEEauthorrefmark{4},
Yuqing Wang\IEEEauthorrefmark{5},Yuan-Fang Wang\IEEEauthorrefmark{3}and Weining Shen\IEEEauthorrefmark{2}}
\IEEEauthorblockA{\IEEEauthorrefmark{2}\textit{Department of Statistics, University of California, Irvine, CA, USA}}
\IEEEauthorblockA{\IEEEauthorrefmark{3}\textit{Department of Computer Science, University of California, Santa Barbara, CA, USA}
\IEEEauthorblockA{\IEEEauthorrefmark{4}\textit{Meta Platforms, Inc., CA, USA}}
\IEEEauthorblockA{\IEEEauthorrefmark{5}\textit{Stanford Univeristy, CA, USA}}    
Emails: \{xiah6, weinings\}@uci.edu, \{rhystracy, yfwang\}@cs.ucsb.edu,} yunzhao20@meta.com, ywang216@stanford.edu

}

\maketitle
\thispagestyle{plain}
\begin{abstract}
This paper presents PathFinder and PathFinderPlus, two novel end-to-end computer vision frameworks designed specifically for advanced setting strategy classification in volleyball matches from a single camera view. Our frameworks combine setting ball trajectory recognition with a novel set trajectory classifier to generate comprehensive and advanced statistical data. This approach offers a fresh perspective for in-game analysis and surpasses the current level of granularity in volleyball statistics. In comparison to existing methods used in our baseline PathFinder framework, our proposed ball trajectory detection methodology in PathFinderPlus exhibits superior performance for classifying setting tactics under various game conditions. This robustness is particularly advantageous in handling complex game situations and accommodating different camera angles. Additionally, our study introduces an innovative algorithm for automatic identification of the opposing team’s right-side (opposite) hitter's current row (front or back) during gameplay, providing critical insights for tactical analysis. The successful demonstration of our single-camera system's feasibility and benefits makes high-level technical analysis accessible to volleyball enthusiasts of all skill levels and resource availability. Furthermore, the computational efficiency of our system allows for real-time deployment, enabling in-game strategy analysis and on-the-spot gameplan adjustments. The source code of our framework is publicly available\footnote{\url{https://github.com/volleyIEEE/VolleyStats}}.

\end{abstract}


\section{Introduction}

Volleyball, one of the most popular team sports worldwide, is renowned for its dynamic strategies and complex tactics. Analyzing volleyball requires a detailed approach that considers many factors influencing each play. Over the past decade, there have been significant strides in integrating computing technology with tactical analysis in sports such as basketball \cite{b8}, \cite{b10}, \cite{b11} and soccer \cite{b2}, \cite{b9}, \cite{b12}.
In contrast to basketball and soccer, the incorporation of computational assistance for volleyball tactical analysis is still in its early stages, holding considerable untapped potential. Current approaches to recording volleyball technical statistics, such as service errors, attack points, attack efficiency, and reception efficiency, offer only a limited perspective of the game dynamics. These metrics are often manually recorded during matches and lack comprehensive support for in-game decision-making.

Scoring in a volleyball match primarily revolves around attacking, which mostly occurs at the end of a rally. Teams employ specific tactics and strategies to determine the optimal player, target location, and technique for ball hitting. Thus, understanding a team's setting tactics and distribution during the game is critical for both coaches and players. By understanding these tactics and distributions, players can adapt their blocking and defensive strategies accordingly. 

Advanced technical statistics \cite{b3}, which include detailed setting patterns and tactics, have been proposed to enhance tactics and in-game analysis. These statistics differentiate between front-row and back-row attack setting patterns (where players jump from either in front of or behind the 3-meter line as per volleyball rules\cite{b20}). Front-row and back-row attacks require vastly different defensive strategies to counter. Thus being able to differentiate between a front- and a back-row set is highly important for volleyball analysis. Additionally, with the evolving athletic capabilities of players, the frequency of back-row sets has increased over the years, making it even more relevant to accurately analyze these types of sets.

Despite their utility, these advanced statistics must be manually labelled from post-game videos. Although current stats are relatively straightforward, they are not universally adopted due to the manual input process and the cost of training a recorder. Therefore, the implementation of these advanced setting pattern statistics in real games, especially in non-top professional level games, is nearly impossible as it requires recorders to possess extensive knowledge of volleyball.

Other sports such as basketball have paved the way by successfully incorporating automatic methods to generate advanced stats that fulfill the demands of professional games and expert analysts. Companies including STATS (Sports VU) \cite{b4}, Second Spectrum \cite{b5}, and NBN23 \cite{b6} have commercialized advanced game statistics using multi-camera detection. Given the high cost of multi-camera usage, affordable alternatives have also been introduced \cite{b7}.

In line with basketball, volleyball also needs an automatic method to produce advanced statistics that meet the needs of modern professional coaches and players, aiding them to make on-court decisions. While Chen \cite{b13} and Chakraborty \cite{b14} proposed a framework to automatically extract ball trajectory and classify/detect the setting pattern, their methods do not differentiate between front-row and back-row attacks. This distinction is becoming increasingly critical as modern volleyball accelerates and places more emphasis on three-dimensional attacks. Commercial solutions such as PlayfulVision \cite{b15} provide tactical statistics for top-level volleyball games using multi-view cameras. Some studies also demonstrate the superior performance of multi-view cameras \cite{b16},\cite{b17}, \cite{b18} in ball trajectory extraction. However, it is important to acknowledge that volleyball, compared to football and basketball, receives less investment, particularly in non-professional level matches. Thus, it remains impractical to employ multi-angle cameras for extracting volleyball information in regions with general levels of play. Nevertheless, the popularization and advancement of volleyball remain achievable goals.

To bridge this gap and empower players of average skill and the general public with these advanced volleyball setting tactics and pattern statistics, we propose a setting tactics pattern recognition framework, PathFinder. PathFinder is a low-cost, end-to-end Computer Vision framework that takes raw videos of volleyball rounds as inputs and outputs detected advanced statistics, including set tactic classification. Our PathFinder framework, along with the improved PathFinderPlus framework, serves as a promising foundation for showcasing the current possibilities of automated advanced volleyball analysis, yielding satisfactory results across various video angles and qualities. Notably, our framework is designed to work with a single camera and incorporates back-row attack recognition. This not only promotes a deeper understanding of volleyball but also enhances the overall enjoyment of the sport, ultimately driving the further growth and popularity of volleyball.

Our framework offers three significant advantages. Firstly, it aligns well with current match recording practices, as many coaches save game recordings round-by-round during the match. Our framework is capable of directly analyzing these recordings and generating an advanced version of real-time setting tactics and pattern statistics, aiding coaches in making informed decisions during the game. Secondly, our solution eases the burden on statistic recording personnel, removing the need for a background in volleyball or an understanding of complex data recording tasks. Lastly, our framework exhibits extensive applicability, catering to various levels of play, including university, high school, and other non-professional games. Irrespective of the specific setting, as long as a camera is available, our algorithm can be easily deployed for analysis and statistics generation.

In summary, our contributions are primarily four-fold:

\begin{itemize}
\item We propose the first end-to-end computer vision framework capable of detecting and classifying volleyball setting patterns, including distinguishing between back-row and front-row setting patterns.
\item We introduce PathFinderPlus, a modified version of our PathFinder framework with a novel ball extraction method that improves the performance of our setting pattern classifier by 4\%-5\% under various game conditions.
\item We are the first to propose an algorithm that leverages rally scoring information to determine if the opposing team is in the back-row or front-row, which can be expanded to track all player game rally rotations.

\item Finally, our system's high computational efficiency allows for real-time deployment. This can enable coaches and players to analyze in-game strategies and make on-the-spot adjustments to their game plans based on the statistics generated by our system. In addition, the system's capability for after-game film study could provide an additional tool for teams to review their performance and strategize for future matches.
\end{itemize}

The remainder of this paper is organized as follows. Section \hyperref[sec:related]{2} discusses related work. The formal problem description is in Section \hyperref[sec:problem]{3}. The proposed framework is described in Section \hyperref[sec:method]{4}. Experiments and results are discussed in Section \hyperref[sec:er]{5}. A discussion of future work is presented in Section \hyperref[sec:con]{6}.

\section{related work}
\label{sec:related}
In this section, we review related work on automated setting pattern classification, as well as volleyball trajectory extraction methodologies.

\subsection{Automatic setting pattern detection and classification}
With the modernization of volleyball, athletes' physical fitness is continually improving, leading to increased speed and height in attacks. As the game progresses, acquiring real-time tactical statistics of the opposing team has become more important than ever. Existing statistical methods, however, are inadequate for monitoring the distribution of the opponent's tactics. Volleyball experts constantly emphasize the importance of real-time access to detailed tactical distribution statistics of opponents, as this information enables coaches to analyze potential setting routes used by the opponents in critical moments, allowing them to establish suitable blocking and defensive systems. This is crucial for securing key points and ultimately, winning the game. Therefore, introducing a more advanced method for setting distribution statistics is essential.

In a 2012 article \cite{b13} by Chen, the author proposed a classification concept for setting tactics. However, this classification has become less significant in modern volleyball era because the importance of back-row attacks has significantly increased over the past decade, and the blocking defense formation corresponding to back-row attacks is completely different from front-row attacks. Failure to  distinguish the difference between these setting patterns can result in ineffective defense and negatively impact the team as a whole. Xia \cite{b3} introduced a technical statistical model in his article that better addresses the requirements of volleyball experts, incorporating separate concepts for front and back-row tactics. However, the implementation of their methods for recognizing setting tactics relied on manual labeling, which is challenging to achieve in real-time gameplay. This is because accurately identifying and categorizing each tactic would necessitate the recorder possessing a moderate level of volleyball knowledge, which is often unrealistic for most locations and levels of matches.  

\subsection{Ball extraction methodologies}

To achieve the objective of automatically recognizing and classifying set patterns, the initial step involves extracting the ball trajectory from video footage. Playfulvision \cite{b15}, Takahashi \cite{b16}, Chen \cite{b17}, and Cheng \cite{b18} have introduced methods for trajectory recognition using 3-D multi-view cameras. However, these methods do not align with our goal of providing players at all levels with access to the advanced technical statistics offered by our framework since it is challenging to ensure the availability of multi-view cameras in most games, expecially at lower levels. For trajectory recognition with a single camera, Chen \cite{b13} and Chakraborty \cite{b14}  combined physical methods with traditional computer vision (CV) techniques for trajectory recognition in 2011 and 2012, respectively. However, as volleyball's speed continues to increase, these methods face greater challenges. In 2020, Toporov \cite{b19} proposed a fusion of deep learning and traditional computer vision for trajectory recognition which provides improved performance and speed. We have based our ball extraction methodologies in our PathFinder framework off of Toporov's method \cite{b19}, as well as improved upon it in our PathFinderPlus framework.

\section{Problem Formulation}
\label{sec:problem}
In modern volleyball games, single-camera in-game videos are commonly used to facilitate coaches' technical analysis post-game. These recordings are typically captured from a camera placed behind one of the teams, allowing for a round-by-round documentation of the game for meticulous examination. Since our input data follows a similar round-based structure, we frame our problem formulation accordingly:

\subsection{Inputs}

\begin{itemize}
    \item $G = \{g_1, g_2, ..., g_n\}$: A volleyball game is represented as a series of rallies, where each rally $g_i$ is a sequence of rounds $g_i = \{r_1, r_2, ..., r_m\}$.
    \item $R = \{r_1, r_2, ..., r_m\}$: Each round consists of a series of video frames $r_i = \{v_1, v_2, ..., v_k\}$ depicting the volleyball play during that round.
    \item $B = \{b_1, b_2, ..., b_q\}$: Within each round, a sequence of the ball's positions are detected from the video frames, represented as the trajectory of the ball $B$.
\end{itemize}

\subsection{Outputs}

$T$: a set of volleyball tactics, $T = \{t_1, t_2, ..., t_p\}$, identified for each round in each rally of the volleyball game.

\subsection{Objective}

Our objective is to accurately identify and classify the setting tactics $T$ used in each round of a volleyball game, given the input video frames $V$ in each round and the ball's trajectory $B$. This can be formalized as an optimization problem where the accuracy of our tactic classification is maximized:

\[
\text{maximize: accuracy}(T, T_{\text{detected}}),
\]
where $T_{\text{detected}}$ are the tactics detected by our model for each round, and accuracy is a function that calculates the fraction of correct tactic detections in each round. The exact definition of accuracy can vary but is usually defined as

\[\text{accuracy}(T, T_{\text{detectd}}) = \frac{\text{number of correct detection}}{\text{total number of detection}}.\]

This optimization problem consists of two parts: ball trajectory detection and tactic classification. In order to complete our objective, we must first achieve precise ball detection in the video frames $V$ to extract the ball's trajectory $B$. However, due to the typically poor quality of cameras used in filming volleyball games, even with advanced computer vision models, this task poses significant challenges. After detecting the ball trajectory, our next step is to determine what setting tactic  $T$ was used. Accurate set tactic classification would enable automated detection and labelling of one of the most important advanced volleyball variables analyzed in Xia et al. \cite{b3}.

\section{Method}
\label{sec:method}
In order to automatically classify volleyball setting strategies and patterns from the video, we break the framework down into the following steps: Ball's Trajectory Extraction, Setting Trajectory Extraction, opposite Front-Back-Row Rotation Recognition, and Setting Path Classification. With all these steps combined, we call this framework \emph{PathFinder}.

\begin{figure*}[htbp]
\centering
\includegraphics[totalheight=0.4\textheight ,width =1\textwidth]{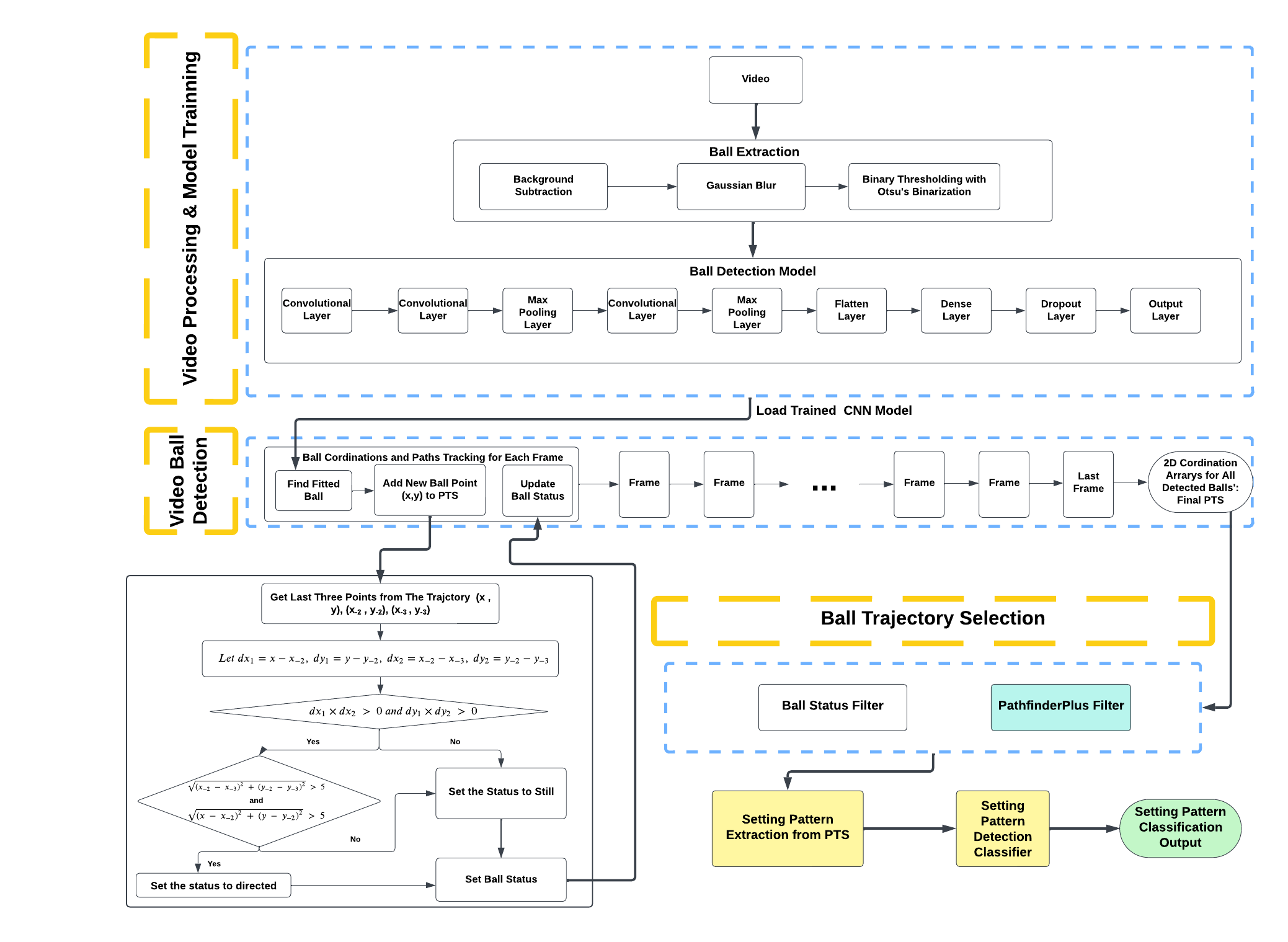}
\caption{Overview of the PathFinder and PathFinderPlus frameworks is presented. The PathFinder framework incorporates Toporov's ball trajectory extraction method \cite{b19} (represented by the uncolored blocks), along with our proposed Setting Pattern Extraction and Setting Pattern Detection Classifier (represented by the yellow blocks) to yield the outcome of the setting pattern tactics classification (represented by the green block). The PathFinderPlus framework enhances the PathFinder by integrating the PathFinderPlus filter (represented by the blue block) to achieve improved setting trajectory extraction results.}
\label{framework}
\end{figure*}

\subsection{ Ball Trajectory Extraction}
\subsubsection{Initial Volleyball Extraction Methodologies}

We initiate our analysis by applying established Computer Vision (CV) techniques, i.e., pre-trained deep neural network models such as YOLOv8 \cite{b22}, for volleyball detection. However, the challenges presented by poor camera quality, noisy backgrounds, and variable ball designs have hindered the success of pre-trained deep neural network models in achieving accurate trajectory prediction, making it largely unfeasible.  Recognizing the limitations of solely relying on these pre-trained deep neural network models, we explore a combination of neural network and traditional CV techniques, specifically tailored for volleyball detection.

Toporov's blended methodology for volleyball extraction\cite{b19} consists of four primary phases: a preprocessing step for ball detection using traditional CV strategies, a Convolutional Neural Network (CNN) for ball tracking (Video processing \& model training in figure ~\ref{framework}), a step for path trajectory detection (Video ball detection in figure ~\ref{framework}), and a filter for ball trajectory selection (ball trajectory selection in figure ~\ref{framework}). The preprocessing phase involves applying Gaussian Blur, Background Subtraction, morphological operations, and Contour Detection to identify potential ball areas from a series of images.

Specifically, Gaussian Blur is employed to minimize high-frequency noise in the images, resulting in smoother images. The Gaussian function in two dimensions is represented as:
\begin{equation}
G(x, y) = \frac{1}{2\pi\sigma^2}e^{-\frac{x^2+y^2}{2\sigma^2}},
\end{equation}
where \(x\) and \(y\) represent the horizontal and vertical distances from the origin,  respectively, and \(\sigma\) denotes the standard deviation of the Gaussian distribution, the selection of which is automated.

Subsequently, a Background Subtractor is utilized to differentiate static and dynamic elements in the video footage. Morphological operations, namely dilation and erosion \cite{b23}, are executed to minimize noise and refine the segmented foreground. Finally, contours are detected and outlined around potential ball regions. These potential ball regions are fed into a pre-trained CNN model for ball classification and tracking. The CNN model, constructed using the Keras library, consists of two Convolutional layers (32 filters for the first and 64 for the second, both with a filter size of 3x3 and ReLU activation), two Max Pooling layers, a Flatten layer, a Fully Connected layer with 64 neurons and ReLU activation, a Dropout layer with a rate of 0.1, and a Softmax output layer with two neurons. The model was compiled using the Stochastic Gradient Descent (SGD) optimizer with a learning rate of 0.01, categorical cross-entropy as the loss function, and accuracy as the performance metric. The model was trained on a dataset of 32x32 RGB images for 50 epochs with a batch size of 32. The trained model weights and architecture were saved for subsequent usage.

In the post-detection phase, this method\cite{b19} utilized a custom-defined Blob class to track and manage the identified balls. This Blob class represents an object being tracked and updated at each frame according to the CNN output. It contains variables such as "id" (a unique identifier for each blob), "pts" (a list that stores the positions of the blob), and status (status of the blob such as "still" or "directed moving").

During tracking, each Blob object maintains a "pts" list, to which the new position of the ball is appended at each frame. This list effectively forms the trajectory of the ball, as it records all the ball positions in a time-sequential manner. This "pts" list is then utilized to predict the ball's next position. It is assumed in the method that the ball's movement is uniform in the short term, hence the next position of the ball can be predicted by fitting a linear model on the most recent positions.

Additionally, the Blob object also holds a "status" property to denote the status of the ball, such as whether it is moving and the direction of its movement. If the ball's position changes between two consecutive frames, the method updates its status to "directed moving". Otherwise, the status remains "still" if the ball's position does not change or if the ball direction is not the same during the last three frames. This process can be denoted as the new added point $(x,y)$, the second last point $(x_{-1}, y_{-1})$, and the third last point $(x_{-2},y_{-2})$ in each 1-d array in the "pts". The method defines $dx1 = x - x_{-1}$, $dx2 = x_{-1} - x_{-2}$. If $dx1 \cdot dx2 > 0$, it indicates that they are in the same direction. The $y$ direction can be checked in a similar way. For the condition of being "still", we check the distance between the newly added point $(x,y)$ with the second last point in each 1-d array in the "pts". If the distance is less than a threshold (e.g., 5 pixels), we consider the ball as "still". This mechanism allows us to detect the ball's movement status while tracking it.

By utilizing the Blob object, we gain access to crucial information about the ball, including its real-time position, movement trajectory, and status. To facilitate this, we created a list of Blob objects, which serves as a container for storing all detected balls within the current frame.  Through an iterative process, we iterate over this list, enabling us to update the position and status of the ball. This procedure ensures the continuous tracking and monitoring of moving balls, providing real-time updates on their positions and status.

Upon completing the tracking process, the final recognized trajectories consist of the "pts" arrays of all Blob objects with the status "directed moving". These trajectories represent the continuous paths followed by all detected balls exhibiting directed movement throughout the series of video frames, hence providing valuable data for further tactical analysis.

However, this original method has its limitations as it only considers whether the most recent movements are in the same direction without effectively filtering false positives. This can lead to inaccuracies as it neglects the overall trend of the ball movement, but instead focusing on localized changes between frames. For instance, certain paths might be mistakenly labeled as "still" because the method was influenced by a few false positives towards the end of the trajectory. These false positives could briefly divert the direction of the ball, causing the method to inaccurately update the ball's status to "still". This scenario higlights the potential issues when the analysis only considers recent movements without studying the overall trend of the ball's movement.

\subsubsection{The PathFinderPlus Ball Detection Approach}
To address the limitations of the original ball extraction method describe above, we propose an improved blended ball detection and trajectory tracking algorithm. The new method is called \emph{PathFinderPlus}, the same PathFinder framework but with superior ball detection and tracking ability. The \emph{PathFinderPlus} ball detection method introduces a filter including two new mathematical functions, namely  \textit{evaluateXDecrease} and \textit{evaluateXIncrease}. 

The \textit{evaluateXDecrease} function is defined as follows: given a series of points $(x_1, y_1), (x_2, y_2), \ldots, (x_n, y_n)$ in the "pts" list, we calculate the differences between consecutive x-coordinates, $dx_i = x_{i+1} - x_i$ for $1 \leq i < n$. We then determine if a majority of these differences are negative. Similarly, the \textit{evaluateXIncrease} function checks if a majority of these differences are positive. Mathematically, this can be interpreted as checking if the derivatives of most of the x-coordinates are negative or positive, respectively.

Furthermore, the \emph{PathFinderPlus} Filter function has been added to the Blob class, which applies the majority decreasing or increasing check on the path of the ball. If a majority of the x-coordinates in the path follow a decreasing or increasing pattern, the ball's path is considered as a valid trajectory.

This global approach is beneficial since it considers the overall trajectory of the ball rather than just the most recent movements. This is particularly useful when there are false positives in the "pts" array, which may temporarily deviate the ball's trajectory but do not affect the overall movement direction. Even if such points appear at the end of the trajectory, our method would still be able to correctly recognize the general movement trend, making it more robust than the original approach.
Therefore, with the \emph{PathFinderPlus} ball detection method, we can more accurately recognize and track the path of moving objects, even when they have some slight changes in their direction due to less-than-ideal ball detection with the poor camera quality. The \emph{PathFinderPlus} ball tracking algorithm also allows us to track and update the status and position of each detected ball in real-time, achieving continuous tracking of moving balls, but with improved accuracy and robustness compared to the original approach.

We are aware that there are other numerical methods for curve fitting and outlier detection \cite{b13,b17} that can be used for ball tracking. For instance, according to the laws of physics, a ball in motion influenced by gravity will follow a parabolic trajectory. By considering the general viewing conditions, where a 3D parabolic curve is projected onto a 2D plane, we can utilize quadratic fitting techniques to predict the future locations of the ball based on its trajectory observed in the game video.

Regarding outlier detection, standard RANSAC \cite{b21} outlier filtering methods can be employed to identify and eliminate outliers in the ball detection process. However, we chose not to utilize such methods for two primary reasons. Firstly, PathFinderPlus's straightforward global approach already demonstrates satisfactory performance in handling outliers. Secondly, since our system is designed for real-time game strategy analysis, we prioritize computational efficiency. Simpler numerical methods tend to excel in this aspect.
In our testing, the PathFinderPlus framework can process a round of data in 16.83 seconds on average on a Macbook Pro equipped with an Apple M1 Pro chip. The average round clip length consists of 152 frames and takes around 6.33 seconds on average. This performance already allows for near real-time analysis. Additionally, considering the breaks between volleyball rallies, which typically last between 15 to 20 seconds (between points scored and the next serve), our framework has even more leeway to achieve real-time analysis during a volleyball match. Furthermore, opting for simpler methods not only supports real-time analysis but also opens up possibilities for IoT (Internet of Things) deployment and reduces hardware requirements, making it more accessible for traditional deployments.

\subsection{Setting Trajectory Extraction}
Both the originally proposed blended volleyball extraction methodology and our PathFinderPlus volleyball detection approaches are only the first step in our proposed PathFinder pipelines. Next we extract the trajectory for the set specifically in order to make a set tactic classification.

In the context of volleyball, a round is typically concluded by a setting action, as it marks the transition to an attack against the opponent's court. Each video segment captures the relevant time span, starting from the pass (or opponent's attack) that initiates a team's possession and ending at the moment when the ball is struck, concluding the possession. The extraction of the ball's path is synchronized with this specific period, ensuring that the trajectory of the set should be the final major trajectory within a round.

In the previous sections, we introduced the "pts" array. Now, in the context of the 'Setting Trajectory Extraction,' we will apply additional processing to this "pts" array to extract the valid setting trajectory for analysis. Mathematically, a "pts" array can be represented as a 2D array $PTS = [A_0, A_1, \ldots, A_{M-1}]$, where each $A_i$ is a 1D array containing two elements that represent the ball's 2D screen positions during the corresponding video frames. The outcome of our framework can be represented as $R = [PTS_0, PTS_1, \ldots, PTS_{N-1}]$, where each $PTS_i$  is a "pts" array. Given that the act of setting predominantly occurs towards the end of a rally, we focus our attention on the terminal ball path within the video segment.

A filtering procedure is then executed to retain only those arrays $PTS_i$ that satisfy $PTS_i \geq 9$. The threshold value of 9 is empirically derived based on the observation that valid setting actions generally yield longer ball path arrays, while shorter arrays more likely indicate false positives or data noise.

We define a function $S(i): \{0, 1, \ldots, M-1\} \rightarrow PTS' \cup \{[0, 0]\}$, where $PTS'$ is the filtered ball path array. $S(i)$ returns $PTS_i$ if $PTS_i \geq 9$. If no such $PTS_i$ exists, $S(i)$ defaults to $[0,0]$. Hence, the final ball path array utilized for trajectory analysis corresponds to $S(M-1)$, signifying the last valid segment in the processed ball path array, or $[0,0]$ in the absence of valid paths.

During this process, we analyze the ball paths in reverse order to extract the setting path more effectively. We remove arrays with a length less than 9. This approach is based on the rationale that, in a volleyball game, the setting action typically constitutes the last substantial trajectory within a round. Any shorter trajectories that follow the setting action are considered noise and are disregarded. In addition, any trajectories preceding the setting action are not relevant to set tactic detection (hence the analysis of the video clip in reverse). To maintain a focus on efficiency and the possibility of fast IoT implementation, we refrain from employing a more complex analysis for trajectory segmentation (i.e., segmenting the trajectory in a round into multiple, sequential quadratic curves, with each quadratic segment representing the ball's trajectory after a contact with a player, the floor, the net, or the net antenna \cite{b13,b17}). 

\subsection{Opposite Front-Back-Row Rotation Recognition}
In modern volleyball, the attack from the back row on the right side of the court, commonly referred to as the "D-ball", is highly significant. The defensive formation required to counter the back-row attack differs greatly from that needed for the front-row attack, specifically the right-side attack, also known as the "Opposite" attack. Therefore, it is crucial to distinguish between the back-row attack (D-ball) and the regular front-row attack (Opposite). To tackle this challenge, we introduce the rotation check procedure.

To understand the rotation check procedure, it is important to discuss the rotation rules in volleyball. The rotation rules in volleyball dictate the positions of the six players on each team during a game. The court is divided into six numbered positions from 1 to 6, as illustrated in Figure~\ref{court}. Positions 1, 5, and 6 are in the back row, while positions 2, 3, and 4 are in the front row. The 3-meter line, indicated by the white dashed line, separates the front row from the back row. This line is crucial in regulating attacks,  as players in the back row have certain restrictions when attacking the ball, such as needing to jump from behind the 3-meter line. Position 1, known as the serving position, is located in the right back area of the court.  

When the receiving team (Team A) wins a rally while the serving team (Team B) is serving, Team A regains the serve and undergoes a rotation. During this rotation, the players on Team A shift their positions in a clockwise direction. Specifically, the player in position 2 moves to position 1 to assume the serving role, the player in position 3 moves to position 2, and so on.  This rotation cycle ensures that players adopt different roles and positions throughout the match.

The rotation check procedure recognizes if the "Opposite" player, the position referring to the player that hits on the right side of the court (e.g., the "opposite" and "d-ball" attacks), is currently in the front or back row. The players' locations will determine if they will be making an "opposite" attack or a "d-ball" attack. We use several notations in this procedure, as outlined in the table~\ref{Opposite Backrow Check Notation table}:

\begin{figure}[htbp]
\centerline{\includegraphics[width=3.5cm, height=3.5cm]{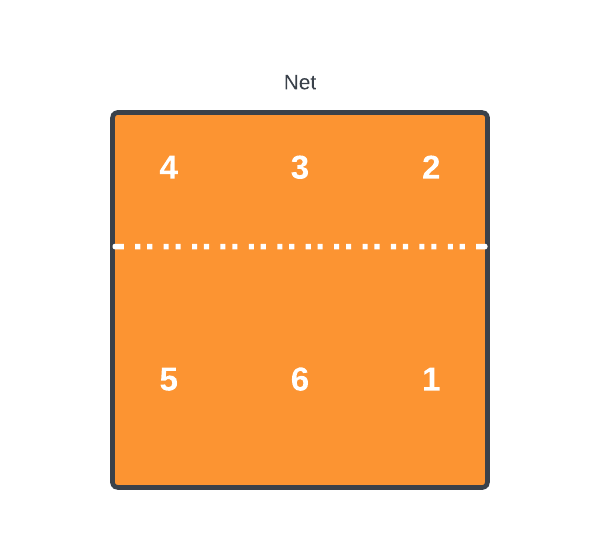}}
\caption{Traditional volleyball court with numbered positions and the 3-meter line.}
\label{court}
\end{figure}

\begin{table}[ht]
\centering
\caption{Opposite Backrow Check Notation table}
\label{Opposite Backrow Check Notation table}
\begin{adjustbox}{width=\columnwidth}
\begin{tabular}{ll}
\hline
\textbf{Symbol} & \textbf{Description} \\
\hline
$\mathit{pos}_A$, $\mathit{pos}_B$ & Initial rotation positions of teams 'A' and 'B' \\
$\mathit{files}$ & Sorted array of files (strings following the pattern "score\_round\_team") \\
$\mathit{score}$, $\mathit{prevScore}$  & The current and previous rally number in the set (resets to 1 with each new set) \\
$\mathit{round}$, $\mathit{prevRound}$ & The current and previous round in the rally(resets to 1 with each new rally) \\
$\mathit{team}$, $\mathit{prevTeam}$ & Team that received the serve first in the current and previous rally \\
$\mathit{BackRow}_A$, $\mathit{BackRow}_B$ & Boolean lists for back-row status of each rotation for teams 'A' and 'B' \\
$\mathit{List}_A$, $\mathit{List}_B$ & Lists of rotation positions for teams 'A' and 'B' \\
$\mathit{opp}_A$, $\mathit{opp}_B$ & Current rotation positions for teams 'A' and 'B' \\
\hline
\end{tabular}
\end{adjustbox}
\end{table}

Algorithm \ref{algo_back} provides details about the implementation of the rotation check procedure. The algorithm begins by initializing the Opposite positions (Lines 5-6), goes through every file and updates the Opposite's position based on the current and previous rallies (Lines 7-18), and finally checks if the Opposite position is in the back row (Lines 19-23). This procedure is essential for distinguishing between front-row and back-row Opposite attacks. As front-row and back-row Opposite attacks have distinct characteristics, the ability to differentiate them can significantly aid in improving the coaches' in-court decision-making.

\begin{algorithm}
\caption{Opposite Back-Row Check}
\label{algo_back}
\begin{algorithmic}[1]
\REQUIRE $\mathit{pos}_A, \mathit{pos}_B, \mathit{files}$
\ENSURE $\mathit{BackRow}_A, \mathit{BackRow}_B$
\STATE \textbf{Procedure} $\mathit{RotationCheck}(\mathit{pos}_A, \mathit{pos}_B, \mathit{files})$
    \STATE $\mathit{List}_A, \mathit{List}_B \leftarrow \{\}, \{\}$
    \STATE $\mathit{opp}_A, \mathit{opp}_B \leftarrow Initial_A, Initial_B$
    \FOR{$i = 0 \to \mathit{len}(\mathit{files})-1$}
        \IF{$i = 0$}
            \STATE $\mathit{opp}_A, \mathit{opp}_B \leftarrow \mathit{pos}_A, \mathit{pos}_B$
        \ELSE
            \STATE $\mathit{score}, \mathit{round}, \mathit{team} \leftarrow \mathit{split}(\mathit{files}[i],'\textunderscore')$
                \STATE $\mathit{prevScore}, \mathit{prevRound}, \mathit{prevTeam} \leftarrow \mathit{split}(\mathit{files}[i-\mathit{int}(\mathit{split}(\mathit{files}[i-1],'\textunderscore')[1])])$
            \IF{$\mathit{team} \neq \mathit{prevTeam}$ \AND $\mathit{score} \neq \mathit{prevScore}$}
                \IF{$\mathit{team} = a$} 
                        \STATE $\mathit{opp}_B \leftarrow ((\mathit{opp}_B - 1) \bmod 6)$ or (6 if $\mathit{opp}_B$ is 0 after mod)
                    
                \ELSE
                    \STATE $\mathit{opp}_A \leftarrow ((\mathit{opp}_A - 1) \bmod 6)$ or (6 if $\mathit{opp}_A$ is 0 after mod)
                \ENDIF
            \ENDIF
        \ENDIF
        \STATE $\mathit{List}_A, \mathit{List}_B \leftarrow \mathit{List}_A \cup \{\mathit{opp}_A\}, \mathit{List}_B \cup \{\mathit{opp}_B\}$
    \ENDFOR
    \FOR{$i = 0 \to \mathit{len}(\mathit{List}_A)-1$}
        \STATE $\mathit{List}_A[i], \mathit{List}_B[i] \leftarrow \mathit{List}_A[i] \in \{1,5,6\}, \mathit{List}_B[i] \in \{1,5,6\}$
    \ENDFOR
    \STATE $\mathit{BackRow}_A, \mathit{BackRow}_B \leftarrow \mathit{List}_A, \mathit{List}_B$
\STATE \textbf{End Procedure}
\end{algorithmic}
\end{algorithm}

\subsection{Setting Path Classification}
The setting path is a crucial element in a volleyball match as it reveals the strategic intentions of a team, providing valuable insights for the coach's decision-making process. In the preceding sections, we have discussed the methodology for detecting the ball's trajectory, extracting the set pattern, and determining the position of the "Opposite" player in the front or back row using a single-camera video. The notations and definitions used in the algorithm are summarized in Table~\ref{set_notation_table}.

To this end, we present an algorithm designed to automatically recognize and classify setting paths based on the positions of the setter and hitter, as well as the trajectory of the ball. The PathFinder algorithm also has the capability to output intermediate-step advanced variables such as setter and hitter contact heights. The detailed set classification process is outlined in Algorithm~\ref{algo_trajectory}. Note that the Coefficients Q, M, S, and C are heuristic values determined through manual analysis by volleyball experts and scaled by the net width in the 2D camera space to accommodate different game scenarios and technical camera angles. In essence, this algorithm examines the maximum height of the set trajectory and the starting and ending locations to determine the set tactic using these specifically crafted heuristics. The two main factors that distinguish different set tactics are the height of the set, which also decides the speed of the set, and the relative location of the hitter with respect to the setter.  These fundamental factors provide a clear framework for a heuristic-based approach to set tactic classification. With the completion of the set classification step, the PathFinder and PathFinderPlus frameworks are finalized.

\begin{table}[ht]
\centering
\caption{Setting Trajectory Analysis Notation Table}
\label{set_notation_table}
\begin{adjustbox}{width=\columnwidth}
\begin{tabular}{ll}
\hline
\textbf{Symbol} & \textbf{Definition} \\
\hline
$B$ & A 2D array representing the ball's trajectory. \\
$LNX$ & The x-coordinate of the left side of the net. \\
$RNX$ & The x-coordinate of the right side of the net. \\
$UNY$ & The y-coordinate of the top of the net. \\
$LNY$ & The y-coordinate of the bottom of the net. \\
$BRA$ & Boolean indicating if there is a back row player ready to spike in team A. \\
$BRB$ & Boolean indicating if there is a back row player ready to spike in team B. \\
$TR$ & String indicating which team is receiving the ball ('A' or 'B'). \\
$P1, P2, P3, P4$ & Points used to divide the court into five equal sections based on x-axis. \\
$NW$ & Net Width. \\
$SP, HP$ & The positions of the setter and hitter along the x-axis, respectively. \\
$HYA$ & The average of the y-coordinates at the highest points of the ball's trajectory. \\
$XD$ & The x-coordinate difference between the hitter and the setter's positions. \\
$T$ & The type of the tactic finally inferred. \\
$Q, M, S, C$ & Heuristic Coefficients for different tactics scaled by net width \\
\hline
\end{tabular}

\end{adjustbox}
\label{tab:notation_table}
\end{table}

\begin{algorithm}
\caption{Analyze Setting Trajectory}
\label{algo_trajectory}
\begin{algorithmic}[1]
\REQUIRE $B, LNX, RNX, UNY, LNY, BRA, BRB, TR$
\ENSURE $T$
\STATE \textbf{Procedure}{Analyze\_Trajectory}{$B$, $LNX$, $RNX$, $UNY$, $LNY$, $BRA$, $BRB$, $TR$}
\STATE $P1, P2, P3, P4 \leftarrow CalculateAreas(LNX, RNX)$
\STATE $NW \leftarrow UNY - LNY$
\STATE $SP, HP \leftarrow mean(x[B[1:3]]), mean(x[B[-3:-1]])$
\STATE $HYA \leftarrow mean(y[sort(B, by=y)[1:5]])$
\STATE $XD \leftarrow HP - SP$
\IF{$TR == 'b'$}
    \IF {$XD > 0$ and $XD \leq \frac{1}{5}(RNX-LNX)$ and $HYA > Q \cdot NW$}
        \STATE $T \leftarrow$ "Quick"
    \ELSIF {$XD > \frac{1}{2}(RNX-LNX)$ and $XD \leq \frac{3}{2}(RNX-LNX)$ and $HP > 1.5 \cdot P1$ and $HP < P4$ and $HYA > M \cdot NW$}
        \STATE $T \leftarrow$ "Thirty-One"
    \ELSIF {$XD < 0$ and $abs(XD) \leq \frac{1}{3}(RNX-LNX)$ and $HYA > Q \cdot NW$}
        \STATE $T \leftarrow$ "Back-One"
    \ELSIF {$SP < P3$ and $SP > P1$ and $HP > P3$ and $HP < P4$ and $HYA > S \cdot NW$}
        \STATE $T \leftarrow$ "Short"
    \ELSIF {$HP > P3 + \frac{1}{2}(P4-P3)$}
        \STATE $T \leftarrow$ "Outside"
    \ELSIF {$HP > P1 + \frac{1}{2}(P2-P1)$ and $HP < P3 + \frac{1}{2}(P4-P3)$ and $HYA < C \cdot NW$}
        \STATE $T \leftarrow$ "Bic"
    \ELSIF {$HP < P1 + \frac{1}{2}(P2-P1)$}
        \IF {$BRB$}
            \STATE $T \leftarrow$ "D-ball"
        \ELSE
            \STATE $T \leftarrow$ "Oppo"
        \ENDIF
    \ELSE
        \STATE $T \leftarrow$ "unknown"
    \ENDIF
\ELSE
    \STATE \emph{Perform the same operations and logic as above for team 'b', but with x directions mirrored for team 'a' on the opposite side of the net}
\ENDIF
\RETURN $T$
\end{algorithmic}
\end{algorithm}

\subsection{Method Summary}
In this section, we have presented a comprehensive methodology for analyzing various aspects of a volleyball game using a single-camera recording. Our PathFinder frameworks are based on innovative techniques and algorithms, and are divided into four primary subsections, each addressing a unique aspect of the volleyball game analysis.

Our method goes beyond simple ball tracking and incorporates strategic patterns and player rotations, which are essential for understanding the game dynamics. This comprehensive approach provides the capability to extract meaningful insights from match recordings, which can significantly aid coaches in improving their tactics and strategies. The algorithms presented in these subsections work in synergy, with the outputs of one procedure serving as inputs to the next. This integrated system allows for a seamless flow of information and analysis, enhancing the overall effectiveness of the framework.


In the next section, we aim to empirically validate our method by demonstrating its effectiveness and reliability through tests conducted on various volleyball match recordings. By providing a comprehensive view of volleyball game analysis, our framework aims to empower coaches, players, and analysts to gain deeper insights, interpret game data more effectively, and ultimately improve team performance.

\section{Experiments and Results}
\label{sec:er}
In this section, we will present the experimental setup and results to validate the effectiveness of our setting strategy classification framework, referred to as "PathFinder".  Furthermore, we will analyze the performance of our proposed improved ball detection methodology in "PathFinderPlus". 

\subsection{Experiments Setup}
\begin{figure}[ht!]
    \centering
    \includegraphics[width=\columnwidth]{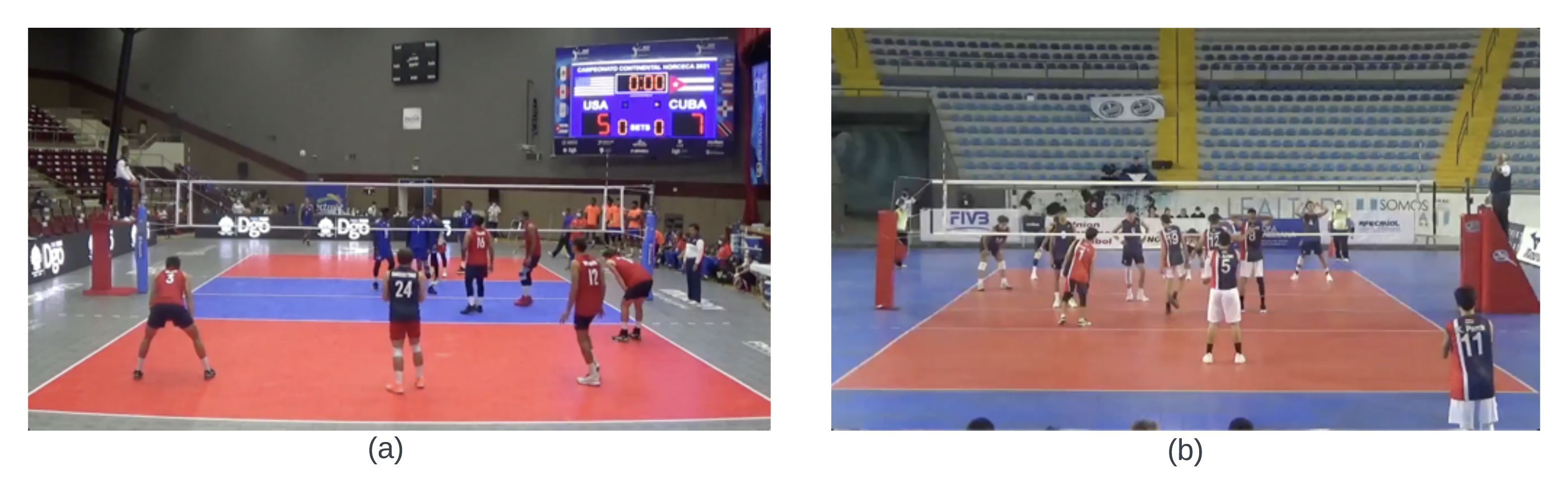}
    \caption{Illustration of the two scenarios used in the experiments. (A) Scenario A: the camera angle is horizontal. (B) Scenario B: the camera angle is slightly downward and the background is noisy.}
    \label{fig:scenarios}
\end{figure}

For our experimentation, we gathered a dataset comprising 537 video clips of volleyball rounds, sourced from 1280 x 720p recordings of national team Men's volleyball match play from 2021-22 (including notable matches such as Cuba vs USA). The experimental setup was designed to simulate two common scenarios encountered in technical video analysis of volleyball matches. In the first scenario (Fig. \ref{fig:scenarios} a), the camera is positioned parallel to the ground, capturing the game from a horizontal perspective. In the second scenario (Fig. \ref{fig:scenarios} b), the camera is positioned at an angle to the ground, and the recorded footage has a relatively complex background. These two scenarios present different challenges for trajectory analysis, player position recognition, and setting path classification. 

\subsubsection{Results}
We compared the performance of our proposed PathFinderPlus with the existing methodology in PathFinder (without our proposed PathFinderPlus filter), whose source code was provided by the original authors, on our setting tactic classifier. The accuracy of the setting tactic classification was assessed across both scenarios. The comparison of the classification results is summarized in Table~\ref{accuracy_comparison}.

\begin{table*}[ht!]
    \centering
    \caption{Comparison of Set Detection Accuracy Under Different Experimental Conditions.
    "Thirty-one" type refers to a setting strategy where the middle blocker hits the ball, with a gap existing between the setter and the hitter. "Quick" involves the middle blocker hitting the ball close to the setter. "Back-one" is similar to "Quick" but here, the middle blocker hits the ball from behind the setter. "Out" refers to the ball being set to an outside hitter position. "Short" refers to a setting strategy where the outside hitter will hit the ball inside, which is closer to the center of the court compared to a normal outside hit. "Bic" refers to a back-row attack where the middle back (player positioned at 6) hits the ball. "Oppo" is used when the opposite (right-side) hitter (the one who traditionally plays across from the setter in the rotation) hits the ball, and the opposite hitter is in the front-row at the time. "D-ball" is used when the opposite hitter hits the ball and they're positioned in the back-row. }
    \label{accuracy_comparison}
    \begin{adjustbox}{width=\textwidth,  height = 1.2cm}
    \begin{tabular}{@{}lcccccccccc@{}}
    \toprule
    \textbf{Experimental Condition} & \textbf{Method} & \multicolumn{9}{c}{\textbf{Accuracy (\%) of Each Set Tactic}}   \\                                                                                                               \\
    & & \textbf{Thirty-one} & \textbf{Out} & \textbf{Oppo} & \textbf{D-ball} & \textbf{Quick} & \textbf{Bic} & \textbf{Back-one} & \textbf{Short} & \textbf{Total} \\
    \midrule
    \multirow{3}{*}{\textbf{Horizontal Camera Angle}} & PathFinder & 36.36\% & 81.67\% & 83.33\% & 69.23\% & 18.75\% & 72.22\% & 25.00\% & 57.14\% & 67.32\% \\ \\
    & PathFinderPlus & 36.36\% & 88.33\% & 83.33\% & 76.92\% & 18.75\% & 77.78\% & 25.00\% & 57.14\% & 71.24\% \\
    \midrule
    \multirow{3}{*}{\textbf{Non-Horizontal Camera Angle and Noisy Background}} & PathFinder & 25.00\% & 60.61\% & 57.14\% & 60.00\% & 18.18\% & 42.86\% & 20.00\% & 33.33\% & 45.89\% \\ \\
    & PathFinderPlus & 37.50\% & 69.70\% & 57.14\% & 60.00\% & 21.21\% & 42.86\% & 20.00\% & 33.33\% & 51.52\% \\ 
    \bottomrule
    \end{tabular}
    \end{adjustbox}
\end{table*}

The results indicate that the PathFinderPlus ball extraction outperforms the existing blended methodology from PathFinder as the first step of our pipeline in both scenarios. Specifically, under the Horizontal Camera Angle condition, the overall accuracy of PathFinderPlus set tactic classification improved over baseline PathFinder from 67.32\% to 71.24\%. Similarly, under the Non-Horizontal Camera Angle and Noisy Background condition, PathFinderPlus demonstrated an increase in overall classification accuracy from 45.89\% to 51.52\%. These results demonstrate the robustness of our pipeline in handling diverse game situations and technical video camera angles. For both ball extraction methods, our framework exhibited a better classification performance under the Horizontal Camera Angle scenario with a simple background than that under the Non-Horizontal Camera Angle and Noisy Background scenario.

\subsection{Case Study: Analyzing the Accuracy Difference}
\begin{figure}[ht!]
    \centering
    \includegraphics[width=\columnwidth]{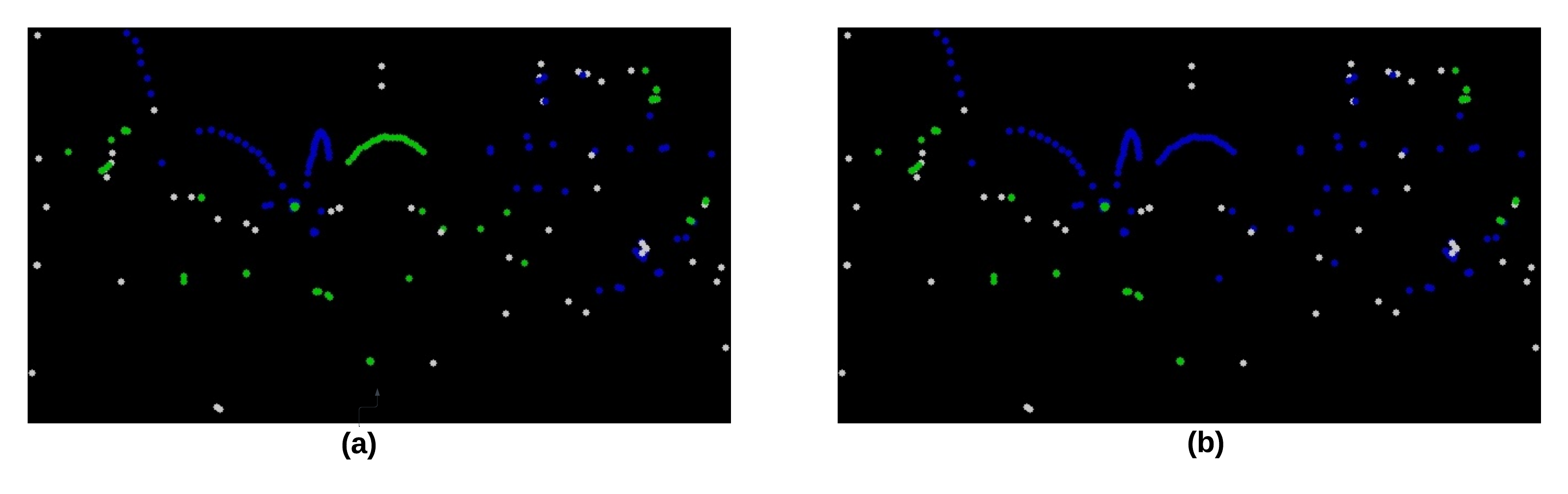}
    \caption{The green dashed line indicates the "still" trajectory status set, i.e., the path is not valid. The blue dashed line indicates a "directed moving"  trajectory status set, i.e., the trajectory is valid. The white dot represents an object mislabeled as a ball (e.g., a players' head), possibly due to visual similarities, but its movement  does not even form a trajectory.}
    \label{path}
\end{figure}

There are two primary factors that contribute to the superior performance of our ball extraction methodology in place of the original blended methodology in our pipeline: 

\begin{itemize}

\item \textbf{Camera Angle and Background Complexity:} Scenario 2 involves a camera angle and more complex backgrounds, leading to instances where the ball might go out of the frame, making it more challenging to track the ball's trajectory. This results in a decrease in accuracy for both methods in Scenario 2 compared to Scenario 1. However, PathFinderPlus manages to mitigate the adverse effect to a certain extent due to its robust design, resulting in higher accuracy under these challenging conditions.

\item\textbf{Frame Analysis Depth:} The original methodology only inspects the last three frames of a setting trajectory, leading to potential misclassifications. For instance, if the last three frames happen to be false positives, the trajectory will be labeled incorrectly. In contrast, our method incorporates a more comprehensive frame analysis, thereby reducing the chances of such misclassification. Figure \ref{path} (a) illustrates a ball trajectory drawn using the original methodology. In this figure,the blue dots represent detected ball locations categorized as "directed moving" and included in the classification pipeline, while the green dots represent detected ball locations labeled as "still" and deemed invalid for classification.  It is evident that the trajectory of the ball is clearly a valid set trajectory, yet it is considered invalid by the original methodology due to false positives at the end of the trajectory. However, Figure \ref{path} (b) demonstrates the same ball trajectory labeled by our PathFinderPlus ball detection algorithm. It successfully recognizes this set trajectory as valid, allowing it to be used for classification later in the pipeline. This showcases the improved accuracy and reliability of our PathFinderPlus methodology in differentiating between valid and invalid ball trajectories.

\end{itemize}

We also note that "Thirty-one", "Back-one", and "Quick" perform relatively worse than other setting tactics. The majority of challenges lie in:

\begin{itemize}
\item \textbf{Challenges of ball detection:} Since all three tactics are for middle blockers, who mostly have a lower setting height and shorter setting distance than other positions, it is difficult for the camera to capture the ball.
\item \textbf{Challenges of misclassification:} In high-level volleyball games, the setting height for "Bic", where players hit from the middle back-row, is similar, albeit with slight differences from these three setting tactics. This similarity poses a challenge in distinguishing them, leading to misclassification.
\end{itemize}

In summary, our proposed PathFinderPlus ball detection methodology outperforms the existing ball detection method when used in our set classification framework for volleyball game analysis under varied conditions due to its comprehensive video frame analysis and robust design. Beyond classification, PathFinderPlus also enables the extraction of advanced statistical data from raw volleyball match footage, offering deeper insights for in-game analysis. Our method proves its effectiveness in handling complex game situations and diverse camera angles, making it a valuable tool for coaches, players, and sports analysts. While both PathFinder and PathFinderPlus show promising results, future work will focus on further enhancing the performance of PathFinderPlus, including the ball detection methodology and the overall framework, under more challenging conditions. Additionally, efforts will be made to improve the classification accuracy of middle blockers' tactics and explore the extension of this methodology to other sports.

\section{Conclusion and Future Work}
\label{sec:con}
In this paper, we introduced and evaluated a novel framework for advanced setting strategy classification in volleyball matches. Our primary contributions are four-fold:

\begin{itemize}

\item \textbf{Automated Advanced Statistics:} With our PathFinder and PathFinderPlus frameworks, we have provided comprehensive advanced statistical data for in-game analysis fully automatically using a single camera. To our knowledge, this level of granularity and automation in advanced volleyball statistics has not been achieved before. Our innovative end-to-end frameworks enable us to take raw volleyball round videos as input and deliver advanced volleyball set tactic classifications as output, empowering coaches with timely and focused advanced setting tactics statistics to assist them in making informed decisions during matches.

\item \textbf{Novel PathFinderPlus Ball Trajectory Detection Methodology:} Our proposed ball trajectory detection method in PathFinderPlus has been shown to outperform the existing methodology in both a horizontal and non-horizontal camera angle scenario with a noisy background on setting tactic classification. This demonstrates the robustness of our system in varied game conditions.

\item \textbf{Opposite Row Identification:} In volleyball, the opposing opposite hitter's row (front or back) significantly affects the team's defensive strategy. We are the first to propose an algorithm to automatically identify the opposite's row during gameplay, providing crucial insights for subsequent tactical analysis.

\item \textbf{Efficient Algorithm Design:}
As mentioned, our aim is to provide this analysis in real time during a game so the coaches and players can dynamically adjust strategies and game plans. Hence, our design is geared toward simplicity with potential for real time and IoT implementation. 
\end{itemize}

Furthermore, our study underscores the feasibility and advantages of a single-camera system. This configuration is not only cost-effective, but also widens the accessibility of high-level technical analysis, making it available to volleyball enthusiasts of varying skill levels and resource availability.

In the future, we envision extending this methodology to track all player rotations throughout the game. We also plan to enhance our system's performance under more challenging conditions and explore its application to other sports. We aim to improve our framework accuracy and promote the enjoyment of volleyball by providing sophisticated analytical tools to a broader audience.

\vspace{12pt}
\end{document}